\pdfoutput=1
\documentclass[11pt]{article}

\usepackage{acl}
\usepackage{times}
\usepackage{latexsym}
\usepackage{booktabs} 
\usepackage{tabularx}
\usepackage{amsmath}
\usepackage[T1]{fontenc}
\usepackage[utf8]{inputenc}

\usepackage{microtype}

\usepackage{inconsolata}

\usepackage{graphicx}
\usepackage{svg}

\title{Lost in the Passage: \\Passage-level In-context Learning Does Not Necessarily Need a "Passage"}
\author{Hao Sun \quad Chenming Tang \quad Gengyang Li \quad Yunfang Wu\thanks{\ \ \ Corresponding author.} \\
National Key Laboratory for Multimedia Information Processing, Peking University \\
MOE Key Laboratory of Computational Linguistics, Peking University\\
School of Computer Science, Peking University \\
\texttt{\{2301213218, wuyf\}@pku.edu.cn} \quad 
\texttt{\{tangchenming, ligengyang\}@stu.pku.edu.cn}
\\
}

\begin{document}
\maketitle
\begin{abstract}
By simply incorporating demonstrations into the context, in-context learning (ICL) enables large language models (LLMs) to yield awesome performance on many tasks. In this study, we focus on passage-level long-context ICL for generation tasks and find that LLMs cannot learn the intrinsic relationship between the demonstration passage and 
the generation output.
We conduct  
experiments 
with different LLMs on two typical generation tasks including single-document question answering and distractor generation, 
demonstrating that even a completely meaningless demonstration passage with 1/4 length 
achieves much better performance than the original full passage. 
Analysis via attention and information flow reveals that LLMs pay little attention to 
passages compared to other components in the prompt and little information flows from the passage to other parts of the demonstration, which further confirms our finding. Additionally, experiments on context compression indicate that compression approaches proven effective on other long-context tasks are not 
suitable for passage-level ICL, since simply using shorter meaningless demonstration passages already achieves competitive performance. 
%We will release our code after the publication of this paper.

% This document is a supplement to the general instructions for *ACL authors. It contains instructions for using the \LaTeX{} style files for ACL conferences.
% The document itself conforms to its own specifications, and is therefore an example of what your manuscript should look like.
% These instructions should be used both for papers submitted for review and for final versions of accepted papers.
\end{abstract}

\section{Introduction}

With recent advancements in prompting strategies, in-context learning (ICL) has become an effective approach to enhancing large language models (LLMs). Instead of updating millions of model parameters, simply incorporating demonstrations into the context enables the model to learn more effectively, achieving better performance than in the zero-shot setting across various tasks. %such as text classification~\cite{min2022rethinkingroledemonstrationsmakes}. 
However, few studies on ICL focus on generation tasks, %while 
and existing research aimed at explaining the underlying mechanism of ICL primarily concentrates on tasks such as sentiment analysis or text classification~\cite{wang2023labelwordsanchorsinformation,min2022rethinkingroledemonstrationsmakes}. 

%Due to the constraints of computational resources and model's context window, 
Different from classification, 
generation tasks, such as question answering (QA) tasks, inherently require long contexts for both query and demonstrations, making it challenging to fit the ICL prompts into model's context window.
%contributing to the scarcity of research in ICL for generation tasks. 
In recent years, with advancements in computing hardware, training data and model architecture, 
%no longer have short context windows of only 512 to 2048 tokens like traditional Transformer based models~\cite{vaswani2023attentionneed}. Instead, 
the context window of LLMs has been expanded to 8K, 32K and even millions of tokens, 
%providing a model basis for ICL in generation tasks and 
allowing researchers to study ICL from the perspective of generation tasks.

However, in this study, we observe a significant phenomenon in passage-level ICL for generation tasks: LLMs cannot capture the intrinsic relationship between the demonstration passage and the corresponding generation target and thus passage-level ICL does not necessarily need a regular well-formed \emph{"Passage"}. 
Specifically, 
%First, 
we adopt Mistral-7B~\cite{jiang2023mistral7b} and Llama2-13B~\cite{touvron2023llama2openfoundation,longlora} models to conduct experiments on two generation tasks: single-document QA and sentence-level distractor generation (DG). For each task, we experiment with randomly generated passages and randomly sampled passages for demonstration. Results show that LLMs are insensitive to demonstration passages in ICL. Even completely meaningless passages
% and generation %targets 
% contents 
in demonstrations do not significantly impact performance. In some cases, they even outperform settings with full-length real passages.

Based on the finding of 
%previous 
prior experiments, we 
%further aim 
% try to 
validate the hypothesis 
%finding 
via attention and information flow analysis.
%analyzing attention scores 
%The attention score experiments consist of two parts: (1) 
First, we compute the attention scores of the first generated token received from different prompt components and those transferred between the passage and other components within each demonstration.
Then, we compute the saliency score matrix and evaluate the significance of information flow between components of demonstration. Results shows that the attention scores LLMs receive from the passage are significantly lower than those from other components, the attention score exchanging within the demonstration is minimal, and only little information flows from passage to generation target. These confirm that LLMs cannot 
%learn 
capture the relationship between the demonstration
passage and %its corresponding 
the generation output. 
%target. %for QA tasks. 
%in the context of generation tasks.

Further, based on the prior experiments, % and interesting findings,  
%Lastly,
we explore context compression for 
%in-context learning of 
ICL. 
%With the development of long-context tasks and models, 
Compressing long contexts into compact texts while minimizing performance degradation has become a crucial approach to handling long-context tasks for effective ICL. 
%This not only reduces computational resource consumption but also %enhances model throughput and reduces 
%speedup inference.
%time. 
Most compression methods deal with long texts  
%focus on other long-context tasks 
rather than on ICL, where the long context contains information relevant to the query, and the main point is to retain key information while filtering out irrelevant content. However, in ICL, the demonstrations themselves do not explicitly contain
%relevant key 
information related to the query. %, and this constitutes the primary distinction between the two scenarios.
%In our work, we seek 
To investigate the effectiveness of compression methods for ICL, we perform compressio experiments on prior passage-level tasks, and the results on both tasks indicate that, under similar compression rates, the existing  compression methods fail to
%significantly 
outperform randomly generated or sampled 
%settings. 
passages.

To sum up, %our contributions are as follows:
0we conduct random perturbation experiments and attention analysis on two ICL tasks. Our results confirm that passage-level ICL does not necessarily need a regular \emph{"Passage"}. Further experiments of context compression show that conventional compression approaches do not provide superior performance to passage-level ICL, since simply using random shorter passages already performs competitively. We hope %anticipate 
this work could inspire further research on the explanation for inner mechanism of ICL and demonstration compression in the passage-level ICL. 

\section{Single-document Question Answering}
\label{qa}
\begin{figure}[ht] % 开始图片环境
\centering % 使图片居中显示
\includegraphics[width=0.45\textwidth]{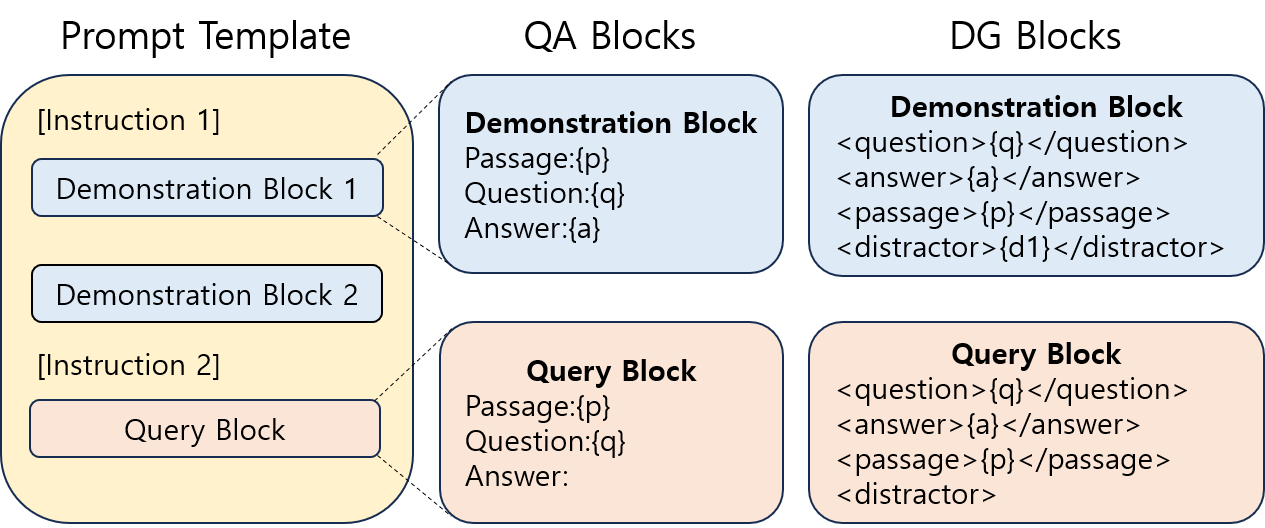} 
\caption{Prompt we use for Single-document QA and Distractor Generation. The left column displays the overall prompt template, with the detailed structures of the demonstration block and query block shown in the other two columns. The middle column presents the blocks used for Single-document QA task, while the right column shows the blocks used for the Distractor Generation task.} % 图片标题
\label{fig/pmt} % 图片标签，用于引用
\end{figure}
%The goal of this section is to verify whether large language models can learn the logical relationships between demonstration passages and their corresponding generation target during in-context learning of passage-level generation tasks.
%The goal of this section is to verify whether LLMs can effectively learn the intrinsic relationships between demonstration passages and their corresponding generation targets during ICL for passage-level generation tasks. To delve deeper into this issue, we 
To examine whether LLMs really comprehend the intrinsic relationship between the passage content and its generation targets during ICL for passage-level generation tasks,
%rather than simply relying on superficial pattern matching or simple imitation to generate results. 
%This study will contribute to a better understanding of the performance and underlying learning mechanisms of large language models in complex text generation tasks. 
%For this purpose, 
we conduct experiments on TriviaQA~\cite{joshi-etal-2017-triviaqa} from Longbench~\cite{bai2024longbenchbilingualmultitaskbenchmark}, which is a single-document QA dataset designed for English reading comprehension. %Specifically, 
We introduce various random perturbations to the demonstrations in the context of ICL and measure the effect on model performance.
%to investigate whether the model effectively learns the intrinsic relationships between the passages and their corresponding QA pairs in the context of ICL.
\subsection{Experimental Setup}
\paragraph{Task Description}
In the QA task for reading comprehension of a single document, each test instance consists of a passage and a question, where the relevant information for the question can be retrieved from the passage. LLMs are required to generate the corresponding answer based on the passage and the question. Evaluation metrics include F1 score, which is used in Longbench, and exact match (EM).
%, considering the average length of the TriviaQA dataset,

\paragraph{LLMs} We use Mistral-7B-Instruct-v0.2~\cite{jiang2023mistral7b} as our primary LLM. It is an instruction-tuned variant of Mistral-7B, which supports a maximum context length of 32K tokens, making it well suited for long-context tasks. Furthermore, we conduct experiments on the LongLoRA fine-tuned variant of the Llama2-13B~\cite{touvron2023llama2openfoundation} model: Llama2-13b-longlora-32k-ft~\cite{longlora}. The LongLoRA fine-tuning extends the context length of Llama2-13B to 32K tokens. All experiments were conducted on a single NVIDIA A100 GPU.

\paragraph{Prompt} Our prompt design adheres to the basic format of TriviaQA and \citet{qu-etal-2024-unsupervised}. Figure \ref{fig/pmt} shows the structure of our prompt and detailed prompt example can be seen in Appendix \ref{app:prompt}. While preserving the original prompt structure of TriviaQA, we incorporate task-specific instructions before both the demonstrations and the query. 

\paragraph{Passage Perturbation} In our systematic perturbation analysis, we mainly employ two methods to perturb the passages in the demonstrations: sampling and generation. For sampling-based perturbation, we randomly sample and reorder tokens from the original passage, ensuring that all the tokens come from the original text. 
%are preserved. 
For generation-based perturbation, we randomly generate a list of numbers as new \texttt{input\_ids} sequences, ensuring that the perturbed results are completely independent. Our goal is to validate the importance of the original passage tokens in the context of ICL through the comparison of these two methods.
% within the token input ids range of the original prompt

Our experiments contain 4-shot and full-shot settings. We apply perturbation ratios of 1/8\footnote{“ratios of 1/8” means randomly generating / sampling tokens with 1/8 length of original passage.}, 1/4, 1/2, 3/4 for all 4-shot settings to the two demonstration passage perturbation methods, evaluating their effects with different LLMs. For full-shot settings, we apply perturbation ratios of 1/8, 1/4, and 3/10~\footnote{3/10 is used because we cannot apply 1/2 due to limited memory.} .
Due to computational resource constraints, full-shot experiments with complete passages are infeasible. For comparison, we include 
%(primarily due to computational resource constraints, full-shot experiments with complete passages were infeasible) 
full-shot setting without any passage (\textit{i.e.}, experiments where the passages in the demonstrations for ICL are entirely removed). For TriviaQA dataset, the full-shot setting means that we use all demonstrations in context for each input from the test data. Each input in the test data contains a different number of demonstrations, ranging from a minimum of 2 to a maximum of 25, with an average of 13.75 and a median of 14.

\subsection{Results and Discussion}
\label{qares}
\begin{table*}[tp]
\renewcommand\arraystretch{1.1}
\centering
\small
\begin{tabular}{l|ccc|ccc}
\bottomrule
\multicolumn{1}{l}{} & \multicolumn{3}{c}{\textbf{Mistral-7B-Instruct-v0.2}} & \multicolumn{3}{c}{\textbf{Llama2-13B-longlora-32k-ft}}\\
\hline
 \textbf{Settings} & \textbf{F1} & \textbf{Exact Match} & \textbf{Avg Prompt Length} & \textbf{F1} & \textbf{Exact Match} & \textbf{Avg Prompt Length}\\
\hline
zero-shot & 47.95 & 27.0 & 580.83 & 74.43 & 66.5 &590.83\\
4-shot + no passage & 73.52 & 63.0 & 669.50 & 71.21 & 67.5 & 669.50\\
4-shot + full passage & 68.52 & 56.0 & 2780.90 & 85.00 & 80.5 & 2780.90\\
\hline
4-shot + generate 1/8 passage & 73.60 & 62.5 & 909.64 & 88.32 & 83.0 & 914.97\\
4-shot + sample 1/8 passage & 71.11 & 59.5 & 925.39 & 86.59 & 82.0 & 926.02\\
4-shot + generate 1/4 passage & \textbf{74.46} & \textbf{63.5} & 1147.31 & \textbf{88.99} & \textbf{84.5} & 1160.67\\
4-shot + sample 1/4 passage & 70.24 & 59.5 & 1193.07 & 86.87 & 82.5 & 1193.48\\
4-shot + generate 1/2 passage & 72.83 & 61.5 & 1627.15 & 88.15 & 83.5 & 1649.72\\
4-shot + sample 1/2 passage & 72.15 & 60.5 & 1721.07 & 88.70 & \textbf{84.5} & 1722.80\\
4-shot + generate 3/4 passage & 71.97 & 61.0 & 2108.30 & 87.30 & 82.5 & 2138.39\\
4-shot + sample 3/4 passage & 68.76 & 58.0 & 2239.56 & 87.89& 84.0 & 2240.77\\
\hline
4-shot + generate question & 63.51 & 49.5 & 2776.65 & 80.70 & 77.0 & 2777.59\\
4-shot + generate answer & 64.18 & 50.0 & 2785.83 & 7.29 & 7.0 & 2785.46\\
\toprule
\end{tabular}
\caption{4-shot results on TriviaQA. The best result in each column is marked in \textbf{bold}.}
\label{tab:tqa_main_4-shot}
\end{table*}
\begin{table*}[tp]
\renewcommand\arraystretch{1.1}
\setlength{\tabcolsep}{4pt}
\centering
\small
\begin{tabular}{l|ccc|ccc}
\bottomrule
  \multicolumn{1}{l}{} & \multicolumn{3}{c}{\textbf{Mistral-7B-Instruct-v0.2}} & \multicolumn{3}{c}{\textbf{Llama2-13B-longlora-32k-ft}}\\
\hline
\textbf{Settings} & \textbf{F1} & \textbf{Exact Match} & \textbf{Avg Prompt Length}  & \textbf{F1} & \textbf{Exact Match} & \textbf{Avg Prompt Length}\\
\hline
%6-shot + full prompt & 69.45 & 56.5 & 3682.33 & 85.58 & 80.0 & 3682.33\\
full-shot + full passage & - & - & 8299.95 & - & - & 8299.95\\
full-shot + no passage & 75.31 & 64.5 & 853.30 & 75.29 & 71.5 & 853.30\\
%\hline
full-shot + generate 1/8 passage & 78.98 & 67.5 & 1701.51 & 88.90 & 84.5 & 1719.44\\
full-shot + sample 1/8 passage & \textbf{79.35} & \textbf{68.0} & 1761.71 & 87.44 & 82.0 &1765.19\\
full-shot + generate 1/4 passage & 78.87 & 67.5 & 2543.09 & 88.51 & 83.5 & 2584.77\\
full-shot + sample 1/4 passage & 78.97 & 67.5 & 2701.92 & 87.06 & 82.0 & 2705.60\\
full-shot + generate 3/10 passage & 77.64 & 65.0 & 2881.74 & 87.51 & 82.5 & 2931.16\\
full-shot + sample 3/10 passage & 77.76 & 66.5 & 3075.57 & \textbf{88.92} & \textbf{84.5} & 3080.37\\
\toprule
\end{tabular}
\caption{Full shot results
%of two models 
on the TriviaQA dataset. The best result in each column is marked in \textbf{bold}.}
\label{tab:tqa_main}
\end{table*}
Experimental results of 4-shot and full-shot settings are presented in Table \ref{tab:tqa_main_4-shot} and \ref{tab:tqa_main} respectively.
The results for both models indicate that LLMs fail to capture the intrinsic relationship between the passages and their corresponding generation targets in the demonstrations of ICL, and they look like \emph{"lost in the passage"}. 

All ICL results show significant improvement compared to the zero-shot setting, indicating that ICL is effective. However, both models demonstrate strong insensitivity to passage perturbations in passage-level ICL, with all results of passage perturbation exceeding the full passage settings.
On Mistral-7B, the F1 and EM scores show an average improvement of 3.37 points and 4.75 points compared to the full passage setting respectively. On Llama2-13B-longlora-32k-ft, the F1 and EM scores achieve an average gain of 2.85 points and 2.81 points respectively.
%Even the full-shot setting with no passage at all achieves better performance than the 6-shot with full prompt setting. which is not as good as other random perturbed settings. 
The 4-shot setting with 1/8 generated passage even reduce the average prompt length from 2780.90 to 909.64 (shortening the length by 67\%) while maintaining the performance, indicating that a significant portion of the context makes no contribution while consuming computational resources.
In contrast, as presented at the bottom of Table \ref{tab:tqa_main_4-shot}, when we try to perturb the question and answer in demonstrations, 
%and this 
it leads to a greater performance degradation than perturbing the passage, indicating that, unlike passages, questions and answers in demonstrations are rather important for passage-level ICL.

\begin{figure*}[t] % 开始图片环境
\centering % 使图片居中显示
\framebox{\includegraphics[width=\textwidth]{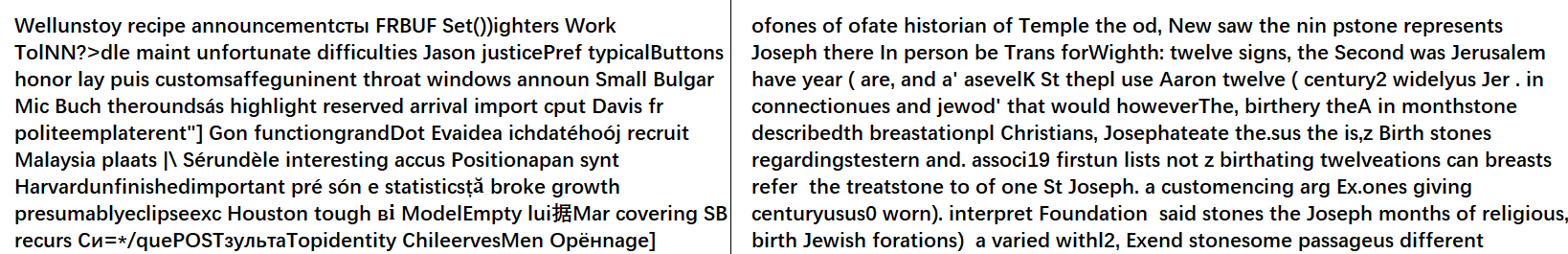} }
\caption{Example of random perturbation passage in demonstration. The passage on the left is random generated passage, and passage on the right is random sampled passage.} % 图片标题
\label{fig/passage} % 图片标签，用于引用
\end{figure*}

Figure \ref{fig/passage} presents two examples of random passage we use in perturbation experiments. The passage on the left presents the random generated passage, which is completely unreadable and meaningless. The random generated words are more diverse, and it even contains words in other languages. On the contrary, the sampled passage on the right is more reasonable than the left passage, whose words are sampled from the original passage.

The comparison between generation and sampling perturbation methods reveals that the settings with randomly generated, completely meaningless passages achieve comparable performance, sometimes even better, to that of sample settings. This indicates that the token sequences sampled from the original passages do not bring improvements. Meanwhile, except for the 4-shot experiment on Mistral-7B, in experiments involving passage content, the performance surpasses that of no passage settings, even when the generated tokens are entirely meaningless. This demonstrates that, in most cases, the presence of content in the passage position is more critical than better content in the passage position. 
%LLMs not only fails to learn the intrinsic relationships between demonstration passage and generation target, but also in fact \emph{"lost in the passage"}, leading to an overall degradation in performance.

We also conduct a detailed ablation study on demonstration selection and Q \& A perturbation. The results can be seen in Appendix \ref{app:qa_abla}.
%The full-shot results of Mistral-7B exceed 4-shot results and we attribute this to the discovery that including more examples provides additional QA pairs, and enables LLMs to better mimic ideal outputs and narrow down the sample space for generation~\cite{min2022rethinkingroledemonstrationsmakes}. 

%In addition, as presented at the bottom of Table \ref{tab:tqa_main_4-shot}, we try to perturb the question \& answer in demonstrations, and this leads to a greater performance degradation than perturbing the passage. 

%In Table \ref{tab:tqa_main}, the comparison between 6-shot full prompt setting and other full-shot settings is in fact unfair. To get a fair comparison, we conduct 4-shot expriments on Llama2-13B-longlora-32k-ft. Table \ref{tab:tqa_abla_shot} presents the results of experiments conducted under the 4-shot setting on the Llama2-32K model, which exhibit similar trends. Notably, perturbing the answer has an even greater impact on the results, further demonstrating that the model fails to learn the intrinsic relationships between the passage and generation target.

%\section{Can Robustness Be Transferred to Other Task?}

\section{Sentence-level Distractor Generation}
\label{dg}
Given that LLMs cannot learn the intrinsic relationship between the demonstration passage and its corresponding demonstration target in single-document QA tasks such as TriviaQA, we further study whether similar trends can be observed in other passage-level ICL scenarios. To this end, we conduct experiments on RACE~\cite{lai-etal-2017-race}, a commonly used dataset sourced from the educational domain and annotated by professional teachers on the DG task, which is more complex than single-document QA. 
%The perturbation methods remain consistent with those applied to the TriviaQA dataset.
%Given that various types of perturbation to the passage have minimal impact on model performance in the question-answering task, is it possible for this phenomenon to occur in other types of tasks?We perturb the dialogue data in the SAMSum dataset in the same manner to study the robustness of different models in this task.

\subsection{Experimental Setup}
\paragraph{Task Description} In the DG task, each instance consists of a document, 
%(corresponding to the passage in single-document QA), 
a question, a correct answer to the question 
%(which can be derived from the document), 
and several distractors designed to mislead the solver. 
%The distractors are extracted from the document and appear to be the plausible answers but are, in fact, counterfactual. 
In our experiments, LLMs are required to generate three distinct distractors.  %Considering the fact that LLMs sometimes generate more than three distractors, we only evaluate the last three distractors in practice. 
Our evaluation metrics include average BLEU 
%(average BLEU) 
and Pairwise BLEU. The former assesses the quality of the generated content, while the latter evaluates the diversity of the generated distractors, with lower values indicating better diversity.
%SAMSum is a dataset designed for the dialogue summarization task, where each input consists of a dialogue text along with a reference summary. Furthermore, the dataset is specifically designed for few-shot learning. Each input instance comprises multiple demonstrations, each demonstration containing a dialogue and its corresponding summary. In our experiments, the model generates a summary for the given dialogue text, and we calculate the similarity between the generated summary and the reference summary.
%In this few-shot dialogue summarization task, the input is divided into two parts: (i) demonstrations, which include the dialogue text and the corresponding summary; (ii) the query part, which contains the dialogue text. Therefore, the goal of this input text is to enable the model to learn the mapping between dialogue and summary presented in the demonstrations, so that the model can generate the correct summary for the dialogue in the query.We conduct experiments with 1-shot, 5-shot, 10-shot and 20-shot settings, applying the same type of perturbation in each group. The group without perturbation served as the baseline for comparison with the other groups, and we observed the changes in ROUGE-L scores before and after perturbation.
Considering that RACE lacks pre-existing input context, we randomly select three sets of examples from the training set and use the same ICL demonstration examples for each test instance. We report the average metrics over three sets of experiments. 

\paragraph{LLMs} Since the prompt length is relatively short, we further include Llama2-13B-Chat alongside the two previously used LLMs, aiming to explore whether LLMs with extended context windows utilize contextual information more effectively. 
%than the original models.

\paragraph{Prompt} Our prompt format aligns with \citet{qu-etal-2024-unsupervised}, whose structure can be seen in Figure \ref{fig/pmt}. We conduct 1, 2, 4, and 8-shot experiments. 

\paragraph{Passage Perturbation} The method for perturbing documents remains consistent with %our previous 
the experiments on TriviaQA. For each few-shot experiment, we configure perturbation ratios of 1/2 and 1/4, with each ratio incorporating both generation-based and sampling-based perturbation methods. 
%Our experimental setup involves three models, each employing a different token manipulation strategy: discarding a portion of tokens, randomly generating tokens of a specified length, and partially replacing some tokens. We compare the performance variations of the models under these different configurations.

%Since this task differs from the QA tasks, the dialogues in the SAMSum dataset are distinct from the passages used in the single-document QA tasks. Although the logical and semantic integrity is preserved even when passages are perturbed in Single-document QA task, dialogues are not only similar to passages, but also questions. Therefore, this task can help us determine whether the robustness of models to perturbations to passage can be transferred to other sections of In-context Learning tasks.

\subsection{Results and Discussion}
\begin{table*}[t]
\renewcommand\arraystretch{1.1}
\setlength{\tabcolsep}{4pt}
\centering
\small
\begin{tabular}{l|l|ccc|ccc|ccc}
\bottomrule
  \multicolumn{2}{l}{} & \multicolumn{3}{c}{\textbf{Llama2-13B-Chat}} &\multicolumn{3}{c}{\textbf{Llama2-13B-longlora-32k-ft}} & \multicolumn{3}{c}{\textbf{Mistral-7B-Instruct-v0.2}} \\
\hline
\textbf{Shot} & \textbf{Settings} & \textbf{AB} & \textbf{PB}($\downarrow$) & \textbf{Avg Length} & \textbf{AB} & \textbf{PB}($\downarrow$) & \textbf{Avg Length} & \textbf{AB} & \textbf{PB}($\downarrow$) & \textbf{Avg Length}\\
\hline
zero-shot & - & 4.32 & 38.52 & 507.69 & 8.06 & 86.18 & 507.69 & 6.46 & 25.28 & 507.69 \\
\hline
1-shot & full & 2.94 & 23.49 & 977.36 & \textbf{4.96} & 37.42 & 977.36 & 4.90 & \textbf{21.41} & 977.36\\
 & no passage & \textbf{4.12} & 26.25 & 546.03 & 3.88 & 45.68 & 546.03 & 4.75 & 22.62 & 546.03\\
 & generate 1/2 & 4.02 & 23.57 & 682.65 & 4.69 & 37.19 & 682.30 & 4.93 & 25.46  & 676.94\\
 & generate 1/4 & 3.72 & 24.87 & 613.56 & 4.58 & 37.11 & 613.58 & 4.79 & 25.13 & 610.46\\
 & sample 1/2 & 3.05 & 23.87 & 764.85 & 4.08 & \textbf{30.40} & 764.85 & 4.83 & 23.15 & 763.47\\
 & sample 1/4 & 3.15 & \textbf{23.05} & 655.87 & 4.48 & 37.25 & 655.87 & \textbf{4.92} & 24.65 & 654.81
\\
\hline
2-shot & full & 4.90 & \textbf{20.78} & 1376.03 & 6.69 & 51.13 & 1376.03 & \textbf{6.08} & \textbf{25.41} & 1376.03\\
 & no passage & 4.46 & 27.73 & 583.03 & 6.39 & 80.48 & 583.03 & 4.47 & 29.23 & 583.03\\
 & generate 1/2 & 5.07 & 22.24 & 835.08 & 6.19 & \textbf{38.58} & 834.83 & 5.24 & 28.05 & 825.36 \\
 & generate 1/4 & \textbf{5.17} & 24.98 & 706.92 & 6.32 & 45.67 & 707.43 & 5.12 & 28.15 & 702.14\\
 & sample 1/2 & 5.00 & 23.59& 978.29 & \textbf{7.61} & 57.64 & 978.27 & 5.37 & 27.48 & 976.76\\
 & sample 1/4 & 4.94 & 25.54 & 782.93 & 7.10 & 59.9 & 782.94 & 4.99 & 28.92 & 781.46\\
\hline
4-shot & full & \textbf{6.01} & \textbf{19.95} & 2052.36 & 5.93 & \textbf{37.35} & 2052.36 & \textbf{6.10} & \textbf{24.41} & 2052.36 \\
 & no passage &4.75 & 28.49 & 666.69 & \textbf{8.58} & 91.68 & 666.69 & 4.49 & 31.54 & 666.69 \\
 & generate 1/2 & 5.35 & 25.03 & 1106.37 & 6.50 & 42.00 & 1105.79 & 4.81 & 29.5 & 1089.84\\
 & generate 1/4 & 5.47 & 27.36 & 882.83 & 7.33 & 53.34 & 882.79 & 4.96 & 30.24 & 873.92\\
 & sample 1/2 & 5.32 & 25.46 & 1344.98 & 6.18 & 42.96 & 1344.99 & 5.46 & 26.65 & 1342.76\\
 & sample 1/4 & 5.29 & 25.87 & 1002.72 & 7.91 & 64.08 & 1002.97 & 5.38 & 28.23 & 1001.16\\
\hline
8-shot & full & - & - & 3759.03 & - & - & 3759.03 & - & - & 3759.03 \\
 & no passage & 4.85 & 29.15 & 797.03 & \textbf{8.12} & 89.98 & 797.03 & 4.74 & 33.01 & 797.03\\
 & generate 1/2 & 5.23 & \textbf{22.32} & 1740.42 & 5.98 & \textbf{38.89} & 1741.50 & 5.10 & 31.01 & 1704.15\\
 & generate 1/4 & \textbf{5.59} & 25.63 & 1261.58 & 6.69 & 47.34 & 1260.66 & 5.30 & 31.31 & 1244.09\\
 & sample 1/2 & 5.40 & 22.73 & 2246.78 & 6.42 & 42.25 & 2246.75 & 5.63 & \textbf{27.64} & 2246.18 \\
 & sample 1/4 & 5.38 & 23.93 & 1506.84 & 7.16 & 53.22 & 1506.97 & \textbf{5.96} & 29.37 & 1505.06\\
\hline
\toprule
\end{tabular}
\caption{Experimental results with three LLMs on RACE dataset with different settings. AB, PB and Avg Length refer to average BLEU, pairwise BLEU, and average prompt length respectively.
%The avg prompt length of 8-shot full passage exceeds the context window of Llama2-13B-Chat, so we neglect this part.
}
\label{tab:dg_main}
\end{table*}
The results are presented in Table \ref{tab:dg_main}, from which we can summarize the following findings.

LLMs exhibit a similar trend in the DG task, showing %complete 
insensitivity to the demonstration passages in ICL. Across different models and numbers of shots, the perturbation settings %with perturbed passages 
achieve comparable and sometimes even better performance than those with full passages, while requiring a much smaller context window.

Compared to previous experiments, the no-passage settings do not always lead to the worst performance. In contrary, 1-shot setting without passage on Llama2-13B-Chat and 4 \& 8-shot settings without passage on longlora model achieve the highest average BLEU among same settings, and they have avg BLEU of 4.12, 8.58, 8.12 respectively. Additionally, one observation is that settings without passage on the two Llama models have the highest pairwise BLEU, %compared with the same model and shot settings, 
with the highest PB reaching 91.68.
%Completely removing the passages from the context has less impact on Avg BLEU scores (sometimes achieving the best performance). However, in most settings, this removal leads to a significant increase in Pairwise BLEU. 
This suggests that retaining some content in the passage position, even if it is entirely meaningless text, can enhance the diversity of the content produced through generation by LLMs. 
%Our experimental results demonstrate that this provides some valuable insight for enriching the content generation capabilities of models.

LLM with context windows extended (Llama2-13B-longlora-32k-ft) achieves better general performance, while is poorer at output diversity and stability. The longlora model has a much higher Avg BLEU score, %in all settings, 
but it suffers from low diversity, with the highest Pairwise BLEU reaching 91.68, which means that the three distractors are almost the same. Moreover, the stability of the model is worse than others, as shown by the drastic fluctuations in both Avg BLEU and Pairwise BLEU. 
%This provides an insight that extending LLMs' context window makes the model better at handling long context and have better performance, but will decrease the model's other abilities such as diversity and stability.

It is noteworthy that the relative orders of Q \& A \& P in the prompt for two tasks are different. As shown in Figure \ref{fig/pmt}, the passage is at the beginning in TriviaQA's prompt, while in that of RACE, it is at the end. However, models show insensitivity to passages in both tasks, indicating that the finding of previous experiments is universal, and the insensitivity of the model to passage is not due to the order of Q \& A \& P, but rather to the model itself.

Detailed results of ablation study on DG can be seen in Appendix \ref{app:abla}.
%This may be due to the tendency of long-context window models to process the overall information of long texts while overlooking local details and tend to generate more tokens, while the base LLama2-13B-Chat can follow instructions. This also explains the larger performance fluctuations observed in these models.

\section{Why Are LLMs Insensitive to Passages?}

In this section, from the aspect of attention and information flow, we provide a deeper confirmation to our hypothesis extracted from former experiments $\mathcal{H}$: \textit{In passage-level ICL, LLMs are in fact unable to capture the intrinsic relationship between demonstration passage and its generation target}.
%In section \ref{atten}, we analyze the average attention scores of the first token received from different components of the prompt, 
%across 5 hidden layers during inference. 
%and this analysis aims to confirm the model's insensitivity to the demonstration passage. Meanwhile in section \ref{re_atten}, we focus on the relative attention scores between the passage and other components of the demonstration, and this directly shows that only little attention is passed between the passage and other parts of the demonstration.

\subsection{Attention Analysis}
\label{atten}
\begin{figure}[tb] % 开始图片环境
\centering % 使图片居中显示
\includegraphics[width=0.45\textwidth]{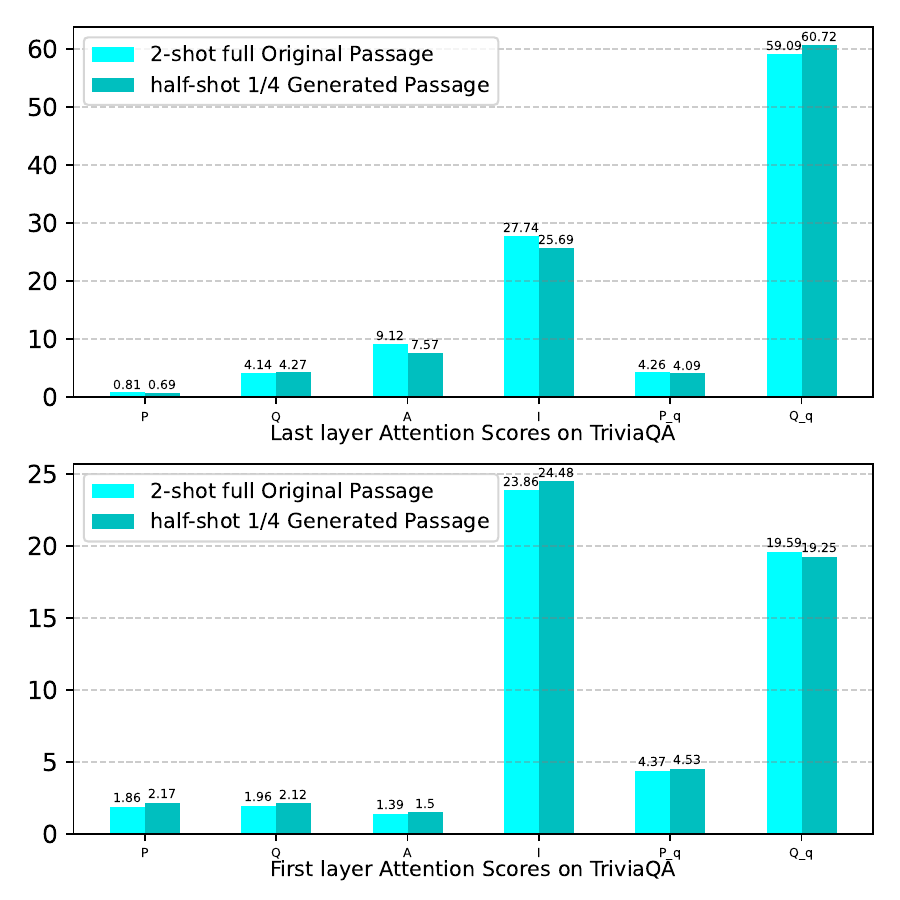} 
\caption{Attention scores of components in prompt on TriviaQA. The horizontal axis index from left to right is \emph{Passage,  Question,  Answer, Instruction, Passage of Query, Question of Query}, respectively.} % 图片标题
\label{fig/qa_atten} % 图片标签，用于引用
\end{figure}
We compute the average attention scores received by the first generated token from different components of the prompt across five hidden layers during inference. This analysis, to some extent, reflects the influence of the prompt's components on the generation. 
%The layers we compute include: the first hidden layer, the 1/4 layer to the first layer in all layers(to edit), 1/2 layer, and the last layer(result of 3/4 layer can be seen in Appendix, which shows similar trend with the last layer*** ), and we believe that this can cover the overall situation during the whole inference process. 
Considering that returning the attention matrix will consume more computing resources than usual, we experiment with two settings for each task. On Trivia QA, we use a 2-shot + full passage prompt and a random half-shot\footnote{"half-shot" means randomly selecting half of the demonstrations.} + generate 1/4 passage. On RACE, we use 2-shot + full passage prompt and 2-shot + random generate 3/4 passage prompt. The partial results on TriviaQA are in Figure \ref{fig/qa_atten} (full results in Appendix \ref{app:fullqa}), 
%and \ref{fig/dg_atten},
and those on RACE are in Appendix \ref{app:dg_atten}.

In the first hidden layer, the attention scores received from different parts of the demonstrations remain approximately equal, indicating that the model does not exhibit a significant preference for any part during the early inference stage. However, in other layers, the attention scores for demonstration passages decrease significantly, falling behind those of other components in the demonstrations. This is consistent across both the full passage and randomly generated passage settings, suggesting that LLMs in fact pay little attention to the demonstration passage. Additionally, an interesting finding is that, apart from the query components, task instruction contributes the most attention to the model. This observation partially explains why modifying instructions can lead to substantial performance changes in certain scenarios.

We also compute the attention scores between different demonstration parts. The results are in Appendix~\ref{app:re_atten}, which also align with our finding.

\subsection{Information Flow Analysis}
\label{saliency}
\begin{figure*}[t] % 开始图片环境
\centering % 使图片居中显示
\includegraphics[width=0.8\textwidth]{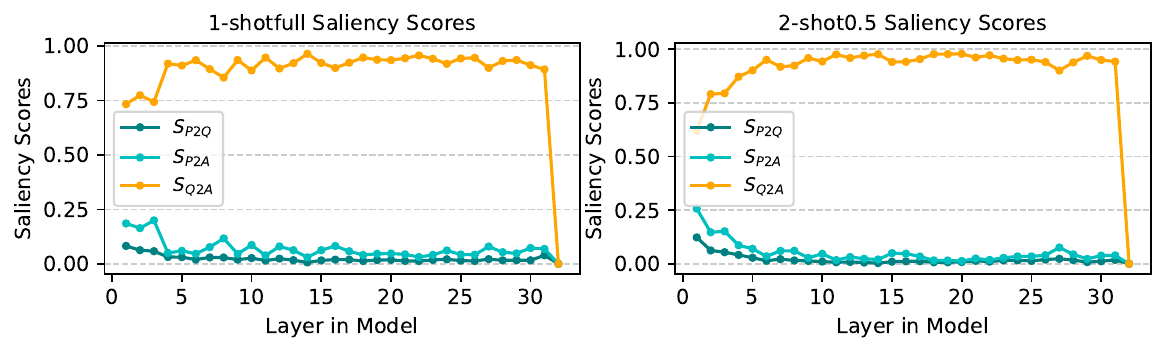} 
\caption{Saliency scores between components of demonstration on TriviaQA.} % 图片标题
\label{fig/saliency_qa} % 图片标签，用于引用
\end{figure*}
We further confirm our hypothesis from the perspective of information flow based on saliency scores~\cite{simonyan2014deepinsideconvolutionalnetworks}. We follow the common practice of computing the saliency score matrix using Taylor expansion~\cite{michel2019sixteenheadsreallybetter,wang2023labelwordsanchorsinformation}. The saliency score matrix at the $l$-th layer $I_l$ is:

\begin{equation}
\label{eq-1}
I_l=\Bigg|\sum_h A_l^h \odot\frac{\partial\mathcal{L}(x)}{\partial A_l^h}\Bigg|,
\end{equation}
where $A_l^h$ represents the attention matrix of the $h$-th head at the $l$-th layer, and $\mathcal{L}(x)$ denotes the loss function (cross-entropy loss for generation tasks in our experiments). The element $I_l(i,j)$ represents the significance of information flow from the the $j$-th token to the $i$-th token. Then, we define several metrics to measure the significance of information flow between components of demonstration. Taking the P2A metric, which measures the information flow between the passage and the answer in demonstration, as an example, we denote $N$ as the number of shots, $p_i^k$ and $a_j^k$ as the $i$-th and $j$-th token of the $k$-th demonstration passage and answer:

\begin{equation}
\label{eq-2}
\begin{aligned}
S_{P2A}^l&=\frac{\sum_{(i,j)\in T_{P2Q}}I_l(i,j)}{|T_{P2Q}|},\\
T_{P2A}&=\{(p^k_i, a^k_j):k \in [1,N]\}.
\end{aligned}
\end{equation}

Calculation of other metrics such as $S_{P2Q}^l$ and $S_{Q2A}^l$ all resemble Equation \ref{eq-2}. Due to computational resource constraints, we conduct experiments only on Mistral-7B for both tasks, with "1-shot full passage" and "2-shot 1/2 generated passage" settings.  The results of TriviaQA are in Figure \ref{fig/saliency_qa} and those of RACE are in Appendix \ref{app:sal}.

In both settings and tasks, the saliency scores between demonstration passage and generation target are relative low. Specifically, across all layers of 
the model on TriviaQA, $S_{P2A}$ is significantly lower than $S_{Q2A}$, and is comparable to $S_{P2Q}$. Both settings exhibit such trend, indicating that settings with full passage cannot benefit the model, in line with our results in Section \ref{qa} and \ref {dg}. This suggests that the information primarily flows from the question to the answer, independent of the passage. Similar trend can be found in RACE experiments, which indicates that information mainly flows from question to the distractors. These demonstrate the model cannot learn the intrinsic relationship between the passage and the generation target, as both tasks require the model to extract information from the passage, yet the model fails to do so.
%\subsection{Attention Score Received by the First Generated Token}

\section{Passage Compression in ICL}
In this section, we explore whether compression algorithms can preserve the most important parts of passages in ICL and achieve better performance than random generation and sampling. 
\subsection{Experimental Setup}
We perform two types of compression methods: retrieval-based and perplexity-based.

In retrieval-based compression~\cite{jiang2024longllmlinguaacceleratingenhancingllms}, we use the question from each ICL demonstration as the retrieval key. After segmenting the passage into sentences, we retrieve the top 5 sentences that are most similar to the question. Additionally, we include a comparison with retrieval re-ranking, where the retrieved sentences are reordered based on the retrieval model's score rather than retaining their original order in the passage. 

For perplexity-based compression, we employ LLMLingua~\cite{jiang2023llmlinguacompressingpromptsaccelerated} and LongLLMlingua~\cite{jiang2024longllmlinguaacceleratingenhancingllms}, two methods exhibiting strong performance on multiple long-document tasks and have been proven to effectively preserve key information. However, in passage-level ICL, the demonstration passages do not directly relate to the query. Our focus is on whether shorter, potentially better passages can help LLMs capture the intrinsic relationship between the passage and its corresponding generation target.

\subsection{Results and Analysis}
Table \ref{tab:comp_qa} shows the results of our compression experiments using Mistral-7B-Instruct-v0.2 on TriviaQA. % for single-document QA task. 
The results indicate that the performance of all compression methods is inferior to that of random generation and sampling. 
%The performance of retrieval-based compression methods is comparable to that of the 6-shot full-passage setting.
The Lingua series methods outperform retrieval-based compression methods, but their performance is still 7 points lower than the average performance of random perturbation experiments. Furthermore, whether re-ranking or randomly selecting demonstrations has a minimal impact on performance, indicating that, in ICL of single-document QA tasks, the presence of content in the passage position is more critical than better content in the passage position.
\begin{table}[tp]
\renewcommand\arraystretch{1.1}
\centering
\small
\begin{tabular}{l|ccc}
\bottomrule
 \textbf{Method} & \textbf{F1} & \textbf{EM} & \textbf{Avg Length}\\
\hline
Our Best & \textbf{79.35} & \textbf{68.0} & 1761.71\\
Our Avg & 78.60 & 67.0 & 2444.26\\
LLMLingua & \textbf{71.36} & \textbf{59.5} & 2246.68\\
LongLingua & 71.33 & 58.5 & 3508.65\\
BM25 Rerank & 67.73 & 56.5 & 2670.90\\
\quad+Random Half Shot & 68.07 & 58.0 & 1602.99\\
BM25 & 67.80 & 56.5 & 2670.90\\
\quad+Random Half Shot& 68.03 & 58.0 & 1602.99\\
Rouge Rerank & 66.87 & 55.0 & 2332.38\\
\quad+Random Half Shot& 67.42 & 56.5 & 1443.69\\
Rouge & 68.03 & 57.0 & 2332.38\\
\quad+Random Half Shot& 67.41 & 57.5 & 1443.69\\
\toprule
\end{tabular}
\caption{Results of TriviaQA compression experiments with Mistral-7B-Instruct-v0.2. Our Best and Our Avg refer to the best and average results from all random perturbation settings. We mark the results of the best setting and our prior best in \textbf{bold}. }
\label{tab:comp_qa}
\end{table}

Table \ref{tab:comp_dg} presents the results using Llama2-13B-Chat on RACE. %dataset for distractor generation. 
Compared to the previous experiments on TriviaQA, the performance of all compression methods is similar, with little performance fluctuations. Notably, although the performance of compression methods in the 4-shot and 8-shot settings is slightly higher than that of random perturbation experiments (improving by approximately 0.5 points), we consider this marginal performance gain insufficient to conclude that compression algorithms allow LLMs to capture the intrinsic relationship between passages and generation targets.%, based on prior ablation studies.
\begin{table}[t]
\renewcommand\arraystretch{1.1}
\setlength{\tabcolsep}{4pt}
\centering
\small
\begin{tabular}{l|l|ccc}
\bottomrule
%   \multicolumn{2}{l}{} &\multicolumn{3}{c}{\textbf{Compress Method}} \\
% \hline
\textbf{Shot} & \textbf{Settings} & \textbf{AB} & \textbf{PB}($\downarrow$) & \textbf{Avg Length}\\
\hline
1-shot & Our Best & \textbf{4.12} & 26.25 & 546.03\\
 & Our Avg & 3.61 & 24.32 & 652.59\\
 & Rouge Rerank & \textbf{3.45} & 28.52 & 646.69\\
 & BM25 Rerank & 3.10 & 25.24 & 651.36\\
 & Rouge & 3.44 & 27.24 & 646.69\\
 & BM25 & 3.14 & 24.67 & 651.36\\
 & LLMLingua & 2.97 & 26.80 & 676.03\\
 & LongLingua & 3.32 & \textbf{21.14} & 734.69\\
\hline
2-shot & Our Best & \textbf{5.17} & 24.98 & 706.92\\
 & Our Avg & 4.93 & 24.82 & 777.25\\
 & Rouge Rerank & 5.29 & 24.39 & 753.69\\
 & BM25 Rerank & 5.24 & 24.08 & 768.69\\
 & Rouge & 5.18 & 24.57 & 753.69\\
 & BM25 & \textbf{5.40} & \textbf{23.19} & 768.69\\
 & LLMLingua & 5.23 & 24.60 & 806.69\\
 & LongLingua & 5.28 & 24.75 & 884.03\\
\hline
4-shot & Our Best & \textbf{5.47} & 27.36 & 882.83\\
 & Our Avg & 5.24 & 26.44 & 1000.72\\
 & Rouge Rerank & 5.53 & 24.32 & 990.69\\
 & BM25 Rerank & 5.76 & 23.63 & 1037.36\\
 & Rouge & 5.62 & 24.49 & 990.69\\
 & BM25 & 5.67 & \textbf{22.56} & 1037.36\\
 & LLMLingua & \textbf{5.92} & 23.72 & 1204.03\\
 & LongLingua & 5.52 & 25.55 & 1284.69\\
\hline
8-shot & Our Best & \textbf{5.59} & 25.63 & 1261.58\\
 & Our Avg & 5.29 & 24.75 & 1510.53\\
 & Rouge Rerank & 6.19 & 24.31 & 1433.69\\
 & BM25 Rerank & \textbf{6.43} & 24.86 & 1531.36\\
 & Rouge & 6.19 & 23.89 & 1433.69\\
 & BM25 & 6.31 & 24.52 & 1531.36\\
 & LLMLingua & 5.98 & \textbf{23.48} & 1898.03\\
 & LongLingua & 5.89 & 23.49 & 2098.69\\
\hline
\toprule
\end{tabular}
\caption{Results of RACE compression experiments with Llama2-13B-Chat. 
%"Our Best" refers to the best result from all random perturbation settings under the same shot. Results of best setting and prior best are in \textbf{bold}.
%B4 refers to BLEU-4, R-L refer to Rouge-L, BS refers to BertScore, PB refers to Pairwise BLEU, FS refers to Faithful Score. 
}
\label{tab:comp_dg}
\end{table}

\section{Related Work}
\subsection{How Do LLMs Utilize the Context?}
Numerous previous studies have explored, from various perspectives, how LLMs utilize context and derive certain insights from ICL. From the perspective of context perturbation, \citet{min2022rethinkingroledemonstrationsmakes} proposes that ground truth demonstrations are not essential. Instead, the label space, the distribution of the input text, and the input format play a more important role in ICL. Furthermore, \citet{liu2023lostmiddlelanguagemodels} finds that the position of key information within the context significantly impacts performance, with key information appearing in the middle position leading to worse performance. Another perspective explains the underlying mechanism of ICL, such as implicit gradient descent during ICL~\cite{dai2023gptlearnincontextlanguage,vonoswald2023transformerslearnincontextgradient} and considering label words as anchors in ICL~\cite{wang2023labelwordsanchorsinformation}.
\subsection{Compression Methods for LLMs}
%There is a substantial amount of prior work on context compression for long-context tasks.
In general, prior work on compression methods can be divided into three categories: extractive method, abstractive method, and soft prompt %compression
method. 

The extractive method selects some tokens 
%and text 
from the original context, ensuring that the compressed results are completely derived from the original context. 
Representative works include selective context~\cite{li2023compressingcontextenhanceinference}, LLMLingua~\cite{jiang2023llmlinguacompressingpromptsaccelerated}, LongLLMLingua~\cite{jiang2024longllmlinguaacceleratingenhancingllms}, LLMLingua2~\cite{pan2024llmlingua2datadistillationefficient} and the ReCOMP extractive compressor~\cite{xu2023recompimprovingretrievalaugmentedlms}. 

The abstractive method aims to generate contextual summaries through language models, ensuring the coherence and fluency of the compression results. 
%Abstractive methods 
including ReCOMP abstractive compressor~\cite{xu2023recompimprovingretrievalaugmentedlms}, Nano-Capsulator~\cite{chuang2024learningcompresspromptnatural}, ComPact~\cite{yoon2024compactcompressingretrieveddocuments}, and semantic compression~\cite{fei2023extendingcontextwindowlarge}.

The soft prompt 
%compression
method compresses the natural language context into soft prompts, %or word embedding, 
aiming to aggregate the key information. Representative works include query-guided compressor~\cite{cao2024retainingkeyinformationhigh} and Dodo~\cite{Qin_2024}.

\section{Conclusion}
In this study, we find that LLMs are unable to capture the intrinsic relationship between the passage and its corresponding generation targets in passage-level ICL. Through experiments on single-document QA and sentence-level DG, we find that randomly perturbing the passage in the demonstrations has minimal impact on performance. 
%whereas perturbing other components more closely related to the generation target leads to the opposite effect.
% Based on above experiments, we analyze the saliency scores \& the relative attention scores between the %passage and other components in demonstrations
% components in demonstrations, as well as attention scores of components of the prompt during inference.
Based on above experiments, we conduct attention and information flow analysis.
The results consistently indicate that LLMs are insensitive to passage during inference. Finally, we introduce compression methods and experimentally show that these methods, while performing well in other long-context tasks, do not provide significant advantages in passage-level ICL. All these results shows that passage-level ICL does not necessarily need a regular \emph{"Passage"}. We hope our finding could inspire future work on explaining the inner mechanisms of ICL.

\section*{Limitations}
First, due to resource limitations, we only study open-source LLMs no larger than 13B and the passage-level ICL performance on larger models, especially powerful models that are extremely good at processing very long context or perturbed content, remains under-explored. Second, we focus on traditional ICL paradigm and use a common prompt template only. The performance is not validated under other paradigms such as chain-of-thought \cite{wei2022chain} and different prompt templates. Furthermore, although we have shown that random perturbation can achieve competitive results with shorter context length compared to representative context compression approaches, how to effectively compress the context for passage-level ICL while keeping stable performance is still unclear and requires future exploration. A promising future direction is combining perturbation and compression since they are orthotropic.

\bibliography{custom}

\begin{thebibliography}{25}
\providecommand{\natexlab}[1]{#1}

\bibitem[{Bai et~al.(2024)Bai, Lv, Zhang, Lyu, Tang, Huang, Du, Liu, Zeng, Hou, Dong, Tang, and Li}]{bai2024longbenchbilingualmultitaskbenchmark}
Yushi Bai, Xin Lv, Jiajie Zhang, Hongchang Lyu, Jiankai Tang, Zhidian Huang, Zhengxiao Du, Xiao Liu, Aohan Zeng, Lei Hou, Yuxiao Dong, Jie Tang, and Juanzi Li. 2024.
\newblock \href {https://arxiv.org/abs/2308.14508} {Longbench: A bilingual, multitask benchmark for long context understanding}.
\newblock \emph{Preprint}, arXiv:2308.14508.

\bibitem[{Cao et~al.(2024)Cao, Cao, Lu, Peng, Huang, Cheng, and Su}]{cao2024retainingkeyinformationhigh}
Zhiwei Cao, Qian Cao, Yu~Lu, Ningxin Peng, Luyang Huang, Shanbo Cheng, and Jinsong Su. 2024.
\newblock \href {https://arxiv.org/abs/2406.02376} {Retaining key information under high compression ratios: Query-guided compressor for llms}.
\newblock \emph{Preprint}, arXiv:2406.02376.

\bibitem[{Chen et~al.(2024)Chen, Qian, Tang, Lai, Liu, Han, and Jia}]{longlora}
Yukang Chen, Shengju Qian, Haotian Tang, Xin Lai, Zhijian Liu, Song Han, and Jiaya Jia. 2024.
\newblock Longlora: Efficient fine-tuning of long-context large language models.
\newblock In \emph{The International Conference on Learning Representations (ICLR)}.

\bibitem[{Chuang et~al.(2024)Chuang, Xing, Chang, Liu, Chen, and Hu}]{chuang2024learningcompresspromptnatural}
Yu-Neng Chuang, Tianwei Xing, Chia-Yuan Chang, Zirui Liu, Xun Chen, and Xia Hu. 2024.
\newblock \href {https://arxiv.org/abs/2402.18700} {Learning to compress prompt in natural language formats}.
\newblock \emph{Preprint}, arXiv:2402.18700.

\bibitem[{Dai et~al.(2023)Dai, Sun, Dong, Hao, Ma, Sui, and Wei}]{dai2023gptlearnincontextlanguage}
Damai Dai, Yutao Sun, Li~Dong, Yaru Hao, Shuming Ma, Zhifang Sui, and Furu Wei. 2023.
\newblock \href {https://arxiv.org/abs/2212.10559} {Why can gpt learn in-context? language models implicitly perform gradient descent as meta-optimizers}.
\newblock \emph{Preprint}, arXiv:2212.10559.

\bibitem[{Fei et~al.(2023)Fei, Niu, Zhou, Hou, Bai, Deng, and Han}]{fei2023extendingcontextwindowlarge}
Weizhi Fei, Xueyan Niu, Pingyi Zhou, Lu~Hou, Bo~Bai, Lei Deng, and Wei Han. 2023.
\newblock \href {https://arxiv.org/abs/2312.09571} {Extending context window of large language models via semantic compression}.
\newblock \emph{Preprint}, arXiv:2312.09571.

\bibitem[{Jiang et~al.(2023{\natexlab{a}})Jiang, Sablayrolles, Mensch, Bamford, Chaplot, de~las Casas, Bressand, Lengyel, Lample, Saulnier, Lavaud, Lachaux, Stock, Scao, Lavril, Wang, Lacroix, and Sayed}]{jiang2023mistral7b}
Albert~Q. Jiang, Alexandre Sablayrolles, Arthur Mensch, Chris Bamford, Devendra~Singh Chaplot, Diego de~las Casas, Florian Bressand, Gianna Lengyel, Guillaume Lample, Lucile Saulnier, Lélio~Renard Lavaud, Marie-Anne Lachaux, Pierre Stock, Teven~Le Scao, Thibaut Lavril, Thomas Wang, Timothée Lacroix, and William~El Sayed. 2023{\natexlab{a}}.
\newblock \href {https://arxiv.org/abs/2310.06825} {Mistral 7b}.
\newblock \emph{Preprint}, arXiv:2310.06825.

\bibitem[{Jiang et~al.(2023{\natexlab{b}})Jiang, Wu, Lin, Yang, and Qiu}]{jiang2023llmlinguacompressingpromptsaccelerated}
Huiqiang Jiang, Qianhui Wu, Chin-Yew Lin, Yuqing Yang, and Lili Qiu. 2023{\natexlab{b}}.
\newblock \href {https://arxiv.org/abs/2310.05736} {Llmlingua: Compressing prompts for accelerated inference of large language models}.
\newblock \emph{Preprint}, arXiv:2310.05736.

\bibitem[{Jiang et~al.(2024)Jiang, Wu, Luo, Li, Lin, Yang, and Qiu}]{jiang2024longllmlinguaacceleratingenhancingllms}
Huiqiang Jiang, Qianhui Wu, Xufang Luo, Dongsheng Li, Chin-Yew Lin, Yuqing Yang, and Lili Qiu. 2024.
\newblock \href {https://arxiv.org/abs/2310.06839} {Longllmlingua: Accelerating and enhancing llms in long context scenarios via prompt compression}.
\newblock \emph{Preprint}, arXiv:2310.06839.

\bibitem[{Joshi et~al.(2017)Joshi, Choi, Weld, and Zettlemoyer}]{joshi-etal-2017-triviaqa}
Mandar Joshi, Eunsol Choi, Daniel Weld, and Luke Zettlemoyer. 2017.
\newblock \href {https://doi.org/10.18653/v1/P17-1147} {{T}rivia{QA}: A large scale distantly supervised challenge dataset for reading comprehension}.
\newblock In \emph{Proceedings of the 55th Annual Meeting of the Association for Computational Linguistics (Volume 1: Long Papers)}, pages 1601--1611, Vancouver, Canada. Association for Computational Linguistics.

\bibitem[{Lai et~al.(2017)Lai, Xie, Liu, Yang, and Hovy}]{lai-etal-2017-race}
Guokun Lai, Qizhe Xie, Hanxiao Liu, Yiming Yang, and Eduard Hovy. 2017.
\newblock \href {https://doi.org/10.18653/v1/D17-1082} {{RACE}: Large-scale {R}e{A}ding comprehension dataset from examinations}.
\newblock In \emph{Proceedings of the 2017 Conference on Empirical Methods in Natural Language Processing}, pages 785--794, Copenhagen, Denmark. Association for Computational Linguistics.

\bibitem[{Li et~al.(2023)Li, Dong, Lin, and Guerin}]{li2023compressingcontextenhanceinference}
Yucheng Li, Bo~Dong, Chenghua Lin, and Frank Guerin. 2023.
\newblock \href {https://arxiv.org/abs/2310.06201} {Compressing context to enhance inference efficiency of large language models}.
\newblock \emph{Preprint}, arXiv:2310.06201.

\bibitem[{Liu et~al.(2023)Liu, Lin, Hewitt, Paranjape, Bevilacqua, Petroni, and Liang}]{liu2023lostmiddlelanguagemodels}
Nelson~F. Liu, Kevin Lin, John Hewitt, Ashwin Paranjape, Michele Bevilacqua, Fabio Petroni, and Percy Liang. 2023.
\newblock \href {https://arxiv.org/abs/2307.03172} {Lost in the middle: How language models use long contexts}.
\newblock \emph{Preprint}, arXiv:2307.03172.

\bibitem[{Michel et~al.(2019)Michel, Levy, and Neubig}]{michel2019sixteenheadsreallybetter}
Paul Michel, Omer Levy, and Graham Neubig. 2019.
\newblock \href {https://arxiv.org/abs/1905.10650} {Are sixteen heads really better than one?}
\newblock \emph{Preprint}, arXiv:1905.10650.

\bibitem[{Min et~al.(2022)Min, Lyu, Holtzman, Artetxe, Lewis, Hajishirzi, and Zettlemoyer}]{min2022rethinkingroledemonstrationsmakes}
Sewon Min, Xinxi Lyu, Ari Holtzman, Mikel Artetxe, Mike Lewis, Hannaneh Hajishirzi, and Luke Zettlemoyer. 2022.
\newblock \href {https://arxiv.org/abs/2202.12837} {Rethinking the role of demonstrations: What makes in-context learning work?}
\newblock \emph{Preprint}, arXiv:2202.12837.

\bibitem[{Pan et~al.(2024)Pan, Wu, Jiang, Xia, Luo, Zhang, Lin, Rühle, Yang, Lin, Zhao, Qiu, and Zhang}]{pan2024llmlingua2datadistillationefficient}
Zhuoshi Pan, Qianhui Wu, Huiqiang Jiang, Menglin Xia, Xufang Luo, Jue Zhang, Qingwei Lin, Victor Rühle, Yuqing Yang, Chin-Yew Lin, H.~Vicky Zhao, Lili Qiu, and Dongmei Zhang. 2024.
\newblock \href {https://arxiv.org/abs/2403.12968} {Llmlingua-2: Data distillation for efficient and faithful task-agnostic prompt compression}.
\newblock \emph{Preprint}, arXiv:2403.12968.

\bibitem[{Qin et~al.(2024)Qin, Rosset, Chau, Rao, and Van~Durme}]{Qin_2024}
Guanghui Qin, Corby Rosset, Ethan Chau, Nikhil Rao, and Benjamin Van~Durme. 2024.
\newblock \href {https://doi.org/10.18653/v1/2024.acl-long.536} {Dodo: Dynamic contextual compression for decoder-only lms}.
\newblock In \emph{Proceedings of the 62nd Annual Meeting of the Association for Computational Linguistics (Volume 1: Long Papers)}, page 9961–9975. Association for Computational Linguistics.

\bibitem[{Qu et~al.(2024)Qu, Sun, and Wu}]{qu-etal-2024-unsupervised}
Fanyi Qu, Hao Sun, and Yunfang Wu. 2024.
\newblock \href {https://doi.org/10.18653/v1/2024.findings-acl.47} {Unsupervised distractor generation via large language model distilling and counterfactual contrastive decoding}.
\newblock In \emph{Findings of the Association for Computational Linguistics: ACL 2024}, pages 827--838, Bangkok, Thailand. Association for Computational Linguistics.

\bibitem[{Simonyan et~al.(2014)Simonyan, Vedaldi, and Zisserman}]{simonyan2014deepinsideconvolutionalnetworks}
Karen Simonyan, Andrea Vedaldi, and Andrew Zisserman. 2014.
\newblock \href {https://arxiv.org/abs/1312.6034} {Deep inside convolutional networks: Visualising image classification models and saliency maps}.
\newblock \emph{Preprint}, arXiv:1312.6034.

\bibitem[{Touvron et~al.(2023)Touvron, Martin, Stone, Albert, Almahairi, Babaei, Bashlykov, Batra, Bhargava, Bhosale, Bikel, Blecher, Ferrer, Chen, Cucurull, Esiobu, Fernandes, Fu, Fu, Fuller, Gao, Goswami, Goyal, Hartshorn, Hosseini, Hou, Inan, Kardas, Kerkez, Khabsa, Kloumann, Korenev, Koura, Lachaux, Lavril, Lee, Liskovich, Lu, Mao, Martinet, Mihaylov, Mishra, Molybog, Nie, Poulton, Reizenstein, Rungta, Saladi, Schelten, Silva, Smith, Subramanian, Tan, Tang, Taylor, Williams, Kuan, Xu, Yan, Zarov, Zhang, Fan, Kambadur, Narang, Rodriguez, Stojnic, Edunov, and Scialom}]{touvron2023llama2openfoundation}
Hugo Touvron, Louis Martin, Kevin Stone, Peter Albert, Amjad Almahairi, Yasmine Babaei, Nikolay Bashlykov, Soumya Batra, Prajjwal Bhargava, Shruti Bhosale, Dan Bikel, Lukas Blecher, Cristian~Canton Ferrer, Moya Chen, Guillem Cucurull, David Esiobu, Jude Fernandes, Jeremy Fu, Wenyin Fu, Brian Fuller, Cynthia Gao, Vedanuj Goswami, Naman Goyal, Anthony Hartshorn, Saghar Hosseini, Rui Hou, Hakan Inan, Marcin Kardas, Viktor Kerkez, Madian Khabsa, Isabel Kloumann, Artem Korenev, Punit~Singh Koura, Marie-Anne Lachaux, Thibaut Lavril, Jenya Lee, Diana Liskovich, Yinghai Lu, Yuning Mao, Xavier Martinet, Todor Mihaylov, Pushkar Mishra, Igor Molybog, Yixin Nie, Andrew Poulton, Jeremy Reizenstein, Rashi Rungta, Kalyan Saladi, Alan Schelten, Ruan Silva, Eric~Michael Smith, Ranjan Subramanian, Xiaoqing~Ellen Tan, Binh Tang, Ross Taylor, Adina Williams, Jian~Xiang Kuan, Puxin Xu, Zheng Yan, Iliyan Zarov, Yuchen Zhang, Angela Fan, Melanie Kambadur, Sharan Narang, Aurelien Rodriguez, Robert Stojnic, Sergey Edunov, and Thomas
  Scialom. 2023.
\newblock \href {https://arxiv.org/abs/2307.09288} {Llama 2: Open foundation and fine-tuned chat models}.
\newblock \emph{Preprint}, arXiv:2307.09288.

\bibitem[{von Oswald et~al.(2023)von Oswald, Niklasson, Randazzo, Sacramento, Mordvintsev, Zhmoginov, and Vladymyrov}]{vonoswald2023transformerslearnincontextgradient}
Johannes von Oswald, Eyvind Niklasson, Ettore Randazzo, João Sacramento, Alexander Mordvintsev, Andrey Zhmoginov, and Max Vladymyrov. 2023.
\newblock \href {https://arxiv.org/abs/2212.07677} {Transformers learn in-context by gradient descent}.
\newblock \emph{Preprint}, arXiv:2212.07677.

\bibitem[{Wang et~al.(2023)Wang, Li, Dai, Chen, Zhou, Meng, Zhou, and Sun}]{wang2023labelwordsanchorsinformation}
Lean Wang, Lei Li, Damai Dai, Deli Chen, Hao Zhou, Fandong Meng, Jie Zhou, and Xu~Sun. 2023.
\newblock \href {https://arxiv.org/abs/2305.14160} {Label words are anchors: An information flow perspective for understanding in-context learning}.
\newblock \emph{Preprint}, arXiv:2305.14160.

\bibitem[{Wei et~al.(2022)Wei, Wang, Schuurmans, Bosma, brian ichter, Xia, Chi, Le, and Zhou}]{wei2022chain}
Jason Wei, Xuezhi Wang, Dale Schuurmans, Maarten Bosma, brian ichter, Fei Xia, Ed~H. Chi, Quoc~V Le, and Denny Zhou. 2022.
\newblock \href {https://openreview.net/forum?id=_VjQlMeSB_J} {Chain of thought prompting elicits reasoning in large language models}.
\newblock In \emph{Advances in Neural Information Processing Systems}.

\bibitem[{Xu et~al.(2023)Xu, Shi, and Choi}]{xu2023recompimprovingretrievalaugmentedlms}
Fangyuan Xu, Weijia Shi, and Eunsol Choi. 2023.
\newblock \href {https://arxiv.org/abs/2310.04408} {Recomp: Improving retrieval-augmented lms with compression and selective augmentation}.
\newblock \emph{Preprint}, arXiv:2310.04408.

\bibitem[{Yoon et~al.(2024)Yoon, Lee, Hwang, Jeong, and Kang}]{yoon2024compactcompressingretrieveddocuments}
Chanwoong Yoon, Taewhoo Lee, Hyeon Hwang, Minbyul Jeong, and Jaewoo Kang. 2024.
\newblock \href {https://arxiv.org/abs/2407.09014} {Compact: Compressing retrieved documents actively for question answering}.
\newblock \emph{Preprint}, arXiv:2407.09014.

\end{thebibliography}

\appendix
\section{Prompt Example}
\label{app:prompt}
We design two different prompt formats for the TriviaQA and RACE datasets, as shown in Table \ref{tab:prompt}. The prompts for both tasks consist of the following components: instructions, demonstrations, task description, and the query-related information. However, there are some differences in the prompts for the two tasks. For TriviaQA, since the questions and answers are typically limited to a single line, the different sections of the prompt are separated by only the newline character '\textbackslash n'. In contrast, the RACE dataset features multiple distractors for the same question and several newline characters within the single passage, which makes it difficult to distinguish different parts with only a single '\textbackslash n'. As a result, we decide to choose the '<>' as a more precise and efficient symbol to locate the corresponding content. In addition, the instructions and task descriptions are designed differently for the two different tasks. This tailored design enables both tasks to achieve strong performance.

When we look closely at the prompts for the two tasks, we can see that the instruction in TriviaQA primarily guides the model to focus on answering QA-type tasks. In contrast, the instruction for the RACE dataset requires the model to generate distractors that align with the relationship between the question and answer. At the same time, both tasks require the model to produce answers in a specified format.

\begin{table*}[ht]
\vspace{-0.3em}
\centering
\footnotesize
\begin{tabularx}{\textwidth}{X|X}
\bottomrule
\multicolumn{1}{c|}{TriviaQA} & \multicolumn{1}{c}{RACE} \\
\hline
You are a helpful AI educational assistant that help students in educational field. You are required to generate answer to the question with the given passage. Next I will propose you several examples. \newline
Passage: $D\_ Passage$ \newline
Question: $D\_Question$ \newline
Answer: $D\_Answer$ \newline
Now according to the following document, question, generate answer for the question.
There are some requirements for you: 1. The returned result can be an incomplete sub-sentence because the grammar structure of the question may be incomplete, but if the return result is incomplete, the combined question-result sentence must have complete grammar structure. 2. Do not generate any irrelvant words. \newline
Passage: $Q\_Passage$ \newline
Question: $Q\_ Question$ \newline
Answer: $Q\_Answer$ &
You are a helpful AI educational assistant that help teachers in educational field. You are required to generate three distractors with the given document, question and answer. Distractors are incorrect answers to the question according to the input document, which are opposite to the answers. The three distractors should be returned in three lines and each line should begin with "<result>" and end with "</result>". Next I will propose you several examples.\newline
<question> $D\_Question$ </question> \newline
<answer> $D\_Answer$ </answer> \newline
<document> $D\_Passage$ </document> \newline
<result> $D\_Distractor$ </result> \newline
Now according to the following document, question and answer, generate three distractors. There are some requirements for you: 1. The returned result can be an incomplete sub-sentence because the grammar structure of the question may be incomplete, but if the return result is incomplete, the combined question-result sentence must have complete grammar structure. 2. The three generated results should be returned in three lines. Each line should begin with '<result>' and end with '</result>' The three distractors can be: <result>\newline
<question> $Q\_Question$ </question> \newline
<answer> $Q\_Answer$ </answer> \newline
<document> $Q\_Passage$ </document> \newline
\\
\toprule
\end{tabularx}
\vspace{-1.0em}
\caption{Prompt for TriviaQA and RACE dataset. $D$ refers to components in demonstrations. $Q$ refers to components in query.}
\label{tab:prompt}
\end{table*}
\section{Ablation Study on single-document QA Task}
\label{app:qa_abla}
\begin{table*}[h]
\renewcommand\arraystretch{1.1}
\centering
\small
\begin{tabular}{l|ccc}
\bottomrule
 \textbf{Settings} & \textbf{F1} & \textbf{Exact Match} & \textbf{Avg prompt length}\\
\hline
Half-shot + generate 1/2 passage & 72.23 & \textbf{62.0} & 2351.52\\
Half-shot + generate 1/4 passage & 71.99 & 60.5 & 1528.64\\
Half-shot + generate 1/8 passage & \textbf{74.97} & 63.5 & 1123.49\\
Half-shot + generate 1/8 passage + random question & 69.48 & 56.5 & 1124.42\\
Half-shot + generate 1/8 passage + random answer &69.57 & 55.0 & 1137.71\\
Half-shot + generate 1/8 passage + random question \& answer & 66.68 & 52.0 & 1132.24\\
\toprule
\end{tabular}
\caption{Results of TriviaQA ablation study about question \& answer perturbation on Mistral-7B-Instruct-v0.2}
\label{tab:tqa_abla_random_qa}
\end{table*}
We conduct ablation studies on Mistral-7B. We introduce random demonstration selection, where we randomly select half of the context demonstrations, and random generation of question and answer in demonstrations. Experimental results are presented in Table \ref{tab:tqa_abla_random_qa}. The results show that randomly selecting half of the ICL examples causes a slight decline in performance, which perhaps results from the reduction of QA pairs. However, perturbing the question-answer pairs exhibits a more substantial impact on model performance. This effect becomes particularly pronounced when both components are altered simultaneously, resulting in significantly decreased F1 and EM scores. And this further confirms the finding that instead of capturing the intrinsic relationship from demonstrations, LLMs tend to mimic the generation target and then generate output based on query~\cite{min2022rethinkingroledemonstrationsmakes}.

\section{Ablation Study on DG Task}
\label{app:abla}
We also conduct ablation study on perturbations of the question, answer, and distractor within the context of ICL demonstrations. In previous experiments, each demonstration contains only one question and answer. In the ablation experiments, we incorporate multiple questions, answers, and distractors from the given dataset into the demonstration in a list format, while keeping the query and other components unchanged. Compared to the perturbation of q\& a \& d in section \ref{qares}, a more regular perturbation will present a credible result . By introducing perturbations to the format of  questions, answers, and distractors in demonstrations, we can more clearly observe that perturbing parts more closely related to the generation target has a greater impact on the model than perturbing passages. The experimental results are presented in Table \ref{tab:dg_abla}.

It is observed that this modification leads to a significant performance degradation. Avg BLEU of almost each setting drops below 3.00, while the Pairwise BLEU remains the same trend. Through case studies, we find that the model's outputs mimic the list format in the demonstrations. The mere introduction of a list format for questions, answers, and distractors results in such a substantial change, whereas completely random generation of passages even improves overall performance in some settings. This reveals the model's insensitivity to the content of the passages.

\begin{table}[ht]
\renewcommand\arraystretch{1.1}
\setlength{\tabcolsep}{4pt}
\centering
\small
\begin{tabular}{l|l|ccc}
\bottomrule
  \multicolumn{2}{l}{} &\multicolumn{3}{c}{\textbf{list q\&a\&d}} \\
\hline
\textbf{shot num} & \textbf{Settings} & \textbf{AB} & \textbf{PB}($\downarrow$) & \textbf{Avg length}\\
\hline
1-shot & Our best & \textbf{4.12} & 26.25 & 546.03\\
 & full & 2.27 & \textbf{24.17} & 1018.69\\
 & no passage  & 2.27 & 28.08 & 587.36\\
 & generate 1/2  & \textbf{2.41} & 26.43 & 723.66\\
 & generate 1/4  & 2.31 & 27.39 & 654.98\\
 & sample 1/2   & 2.26 & 26.20 & 806.16\\
 & sample 1/4 & 2.26 & 26.98 & 697.20\\
\hline
2-shot & Our best & \textbf{5.17} & 24.98 & 706.92\\
 & full  & 2.44 & \textbf{20.50} & 1489.03\\
 & no passage  & 2.69 & 26.28 & 696.03\\
 & generate 1/2 & \textbf{2.76} & 24.55 & 948.59\\
 & generate 1/4  & 2.75 & 24.83 & 820.36\\
 & sample 1/2  & 2.61 & 23.90 & 1091.31\\
 & sample 1/4  & \textbf{2.76} & 25.29 & 895.92\\
\hline
4-shot & Our best & \textbf{5.47} & 27.36 & 882.83\\
 & full  & 2.72 & \textbf{23.60} & 2288.03\\
 & no passage  & 2.76 & 26.77 & 902.36\\
 & generate 1/2  & \textbf{2.81} & 28.40 & 1340.86\\
 & generate 1/4  & 2.72 & 28.26 & 1118.43\\
 & sample 1/2  & 2.66 & 27.55 & 1580.63\\
 & sample 1/4  & \textbf{2.81} & 28.03 & 1238.44\\
\hline
8-shot & Our best & \textbf{5.59} & 25.63 & 1261.58\\
 & full & - & - & - \\
 & no passage  & 2.62 & 27.64 & 1317.36\\
 & generate 1/2  & 2.14 & 35.20 & 2260.99\\
 & generate 1/4 & 2.97 & 26.84 & 1782.64\\
 & sample 1/2  & 2.76 & \textbf{24.72} & 2767.14\\
 & sample 1/4  & \textbf{3.16} & 27.12 & 2027.31\\
\hline
\toprule
\end{tabular}
\caption{Ablation study results of Llama2-13B-Chat on RACE dataset. The prior best refers to the best result from all random perturbation settings under the same shot.
%B4 refers to BLEU-4, R-L refer to Rouge-L, BS refers to BertScore, PB refers to Pairwise BLEU, FS refers to Faithful Score. 
}
\label{tab:dg_abla}
\end{table}

%\label{sec:appendix}

\section{Attention Analysis between Passage and Other Components of Demonstration}
\label{app:re_atten}
\begin{figure}[t] % 开始图片环境
\centering % 使图片居中显示
\includegraphics[width=0.5\textwidth]{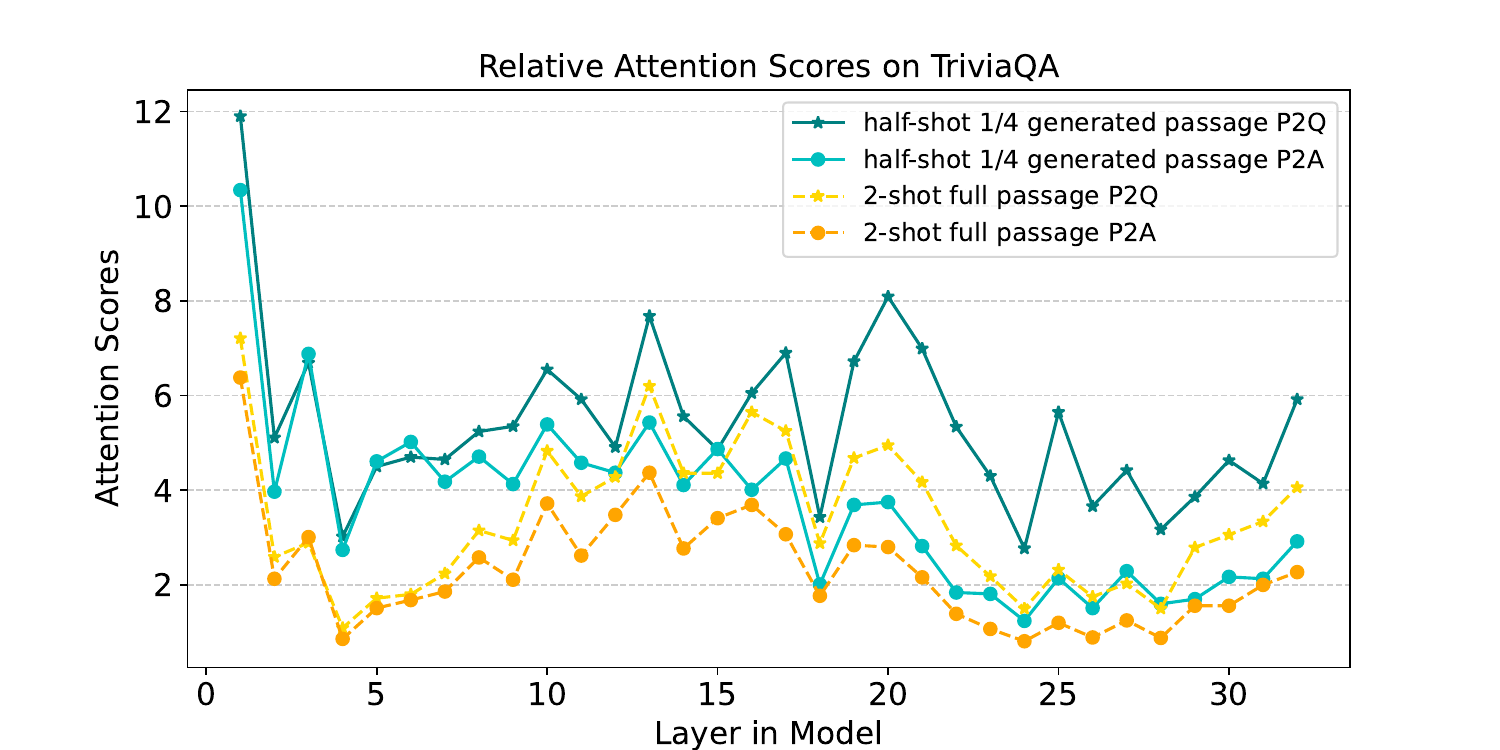} 
\caption{Relative attention scores on TriviaQA with prompts of two settings.
% Layer 1 refers to the first hidden layer in the model.
} % 图片标题
\label{fig/qa_re} % 图片标签，用于引用
\end{figure}

%As mentioned earlier, 
In this section, we directly compute the average attention scores between different parts of the demonstration,
%The overall experimental setup remains consistent with section \ref{atten}, but instead of using the attention scores received by the first generated token, we compute the scores that the demonstration components, 
such as the question, receive from or contribute to the passage, determined by their relative positions in the prompt. Since we only focus on the relative attention scores, we compute the scores on all hidden layers. The results for TriviaQA are presented in Figure \ref{fig/qa_re}, while the results for RACE can be found in Appendix \ref{app:dg_atten}.

As shown in Figure \ref{fig/qa_re}, the attention passed from the demonstration passage to its corresponding answer is lower than that of the question, indicating models' relative insensitivity between the passage and the target. Apart from that, the scores drop after the first layer, and remain at a low level 
%with the most average attention score 
below 6.
%in each setting. 
This aligns with the observation from the previous section, which indicates that the model exhibits no preference for any part of demonstrations during the early stages of inference and pays almost no attention to the passage after the first layer. Results on RACE also reveal this trend.
\subsection{Full Attention Results on TriviaQA}
\label{app:fullqa}
\begin{figure*}[h] % 开始图片环境
\centering % 使图片居中显示
\includegraphics[width=\textwidth]{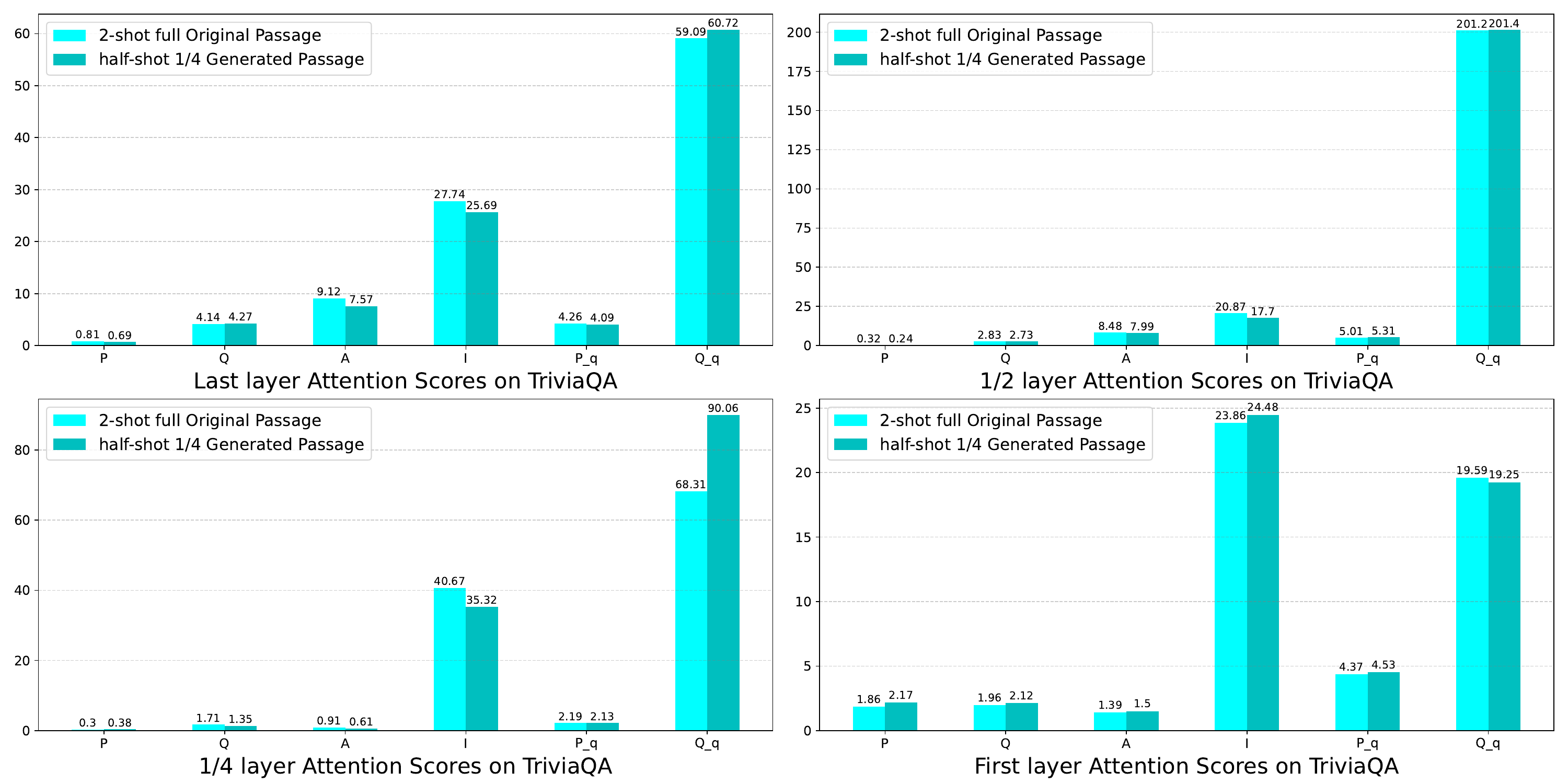} 
\caption{Attention scores of components in prompt on TriviaQA. The horizontal axis index from left to right is \emph{Passage,  Question,  Answer, Instruction, Passage of Query, Question of Query
}, respectively.} % 图片标题
\label{fig/full_qa_atten} % 图片标签，用于引用
\end{figure*}

As previously stated in Section \ref{atten}, the full attention results of TriviaQA can be seen in Figure \ref{fig/full_qa_atten}. As shown in Figure \ref{fig/full_qa_atten} that in other layers including the 1/4 \& 1/2 layer, the attention scores for demonstration passages decrease significantly, falling behind those of other components in the demonstrations, which is consistent across both the full passage and randomly generated passage settings. Figure \ref{fig/full_qa_atten} more comprehensively confirms our finding.
\section{Attention Results on Distractor Generation}
\label{app:dg_atten}
\begin{figure*}[t] % 开始图片环境
\centering % 使图片居中显示
\includegraphics[width=\textwidth]{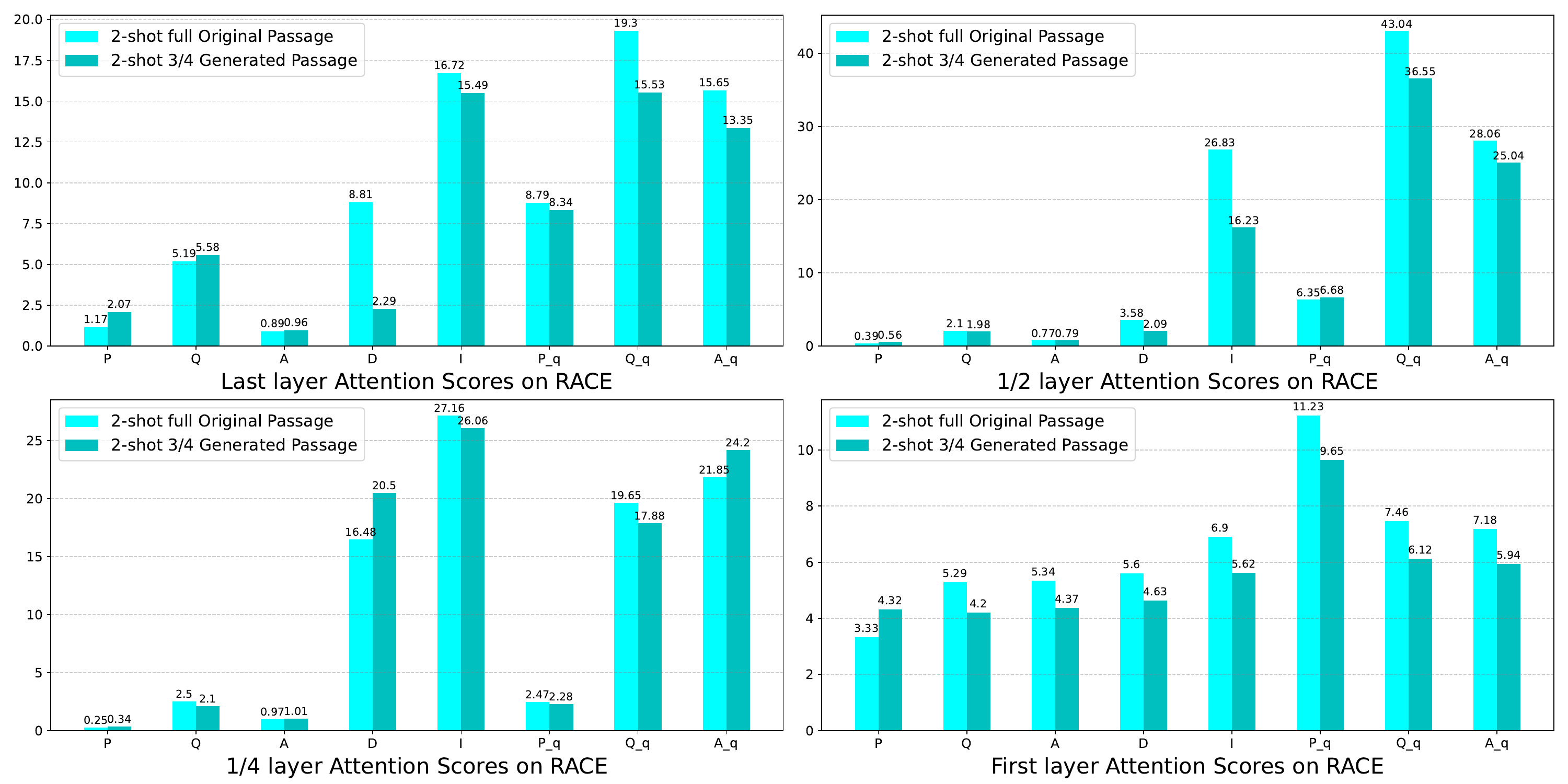} 
\caption{Attention scores of components in prompt on RACE. The horizontal axis index from left to right is \textit{Passage, Question,  Answer, Distractor, Instruction, Passage of Query, Question of Query, Answer of Query}, respectively.} % 图片标题
\label{fig/dg_atten} % 图片标签，用于引用
\end{figure*}

\begin{figure}[ht] % 开始图片环境
\centering % 使图片居中显示
\includegraphics[width=0.5\textwidth]{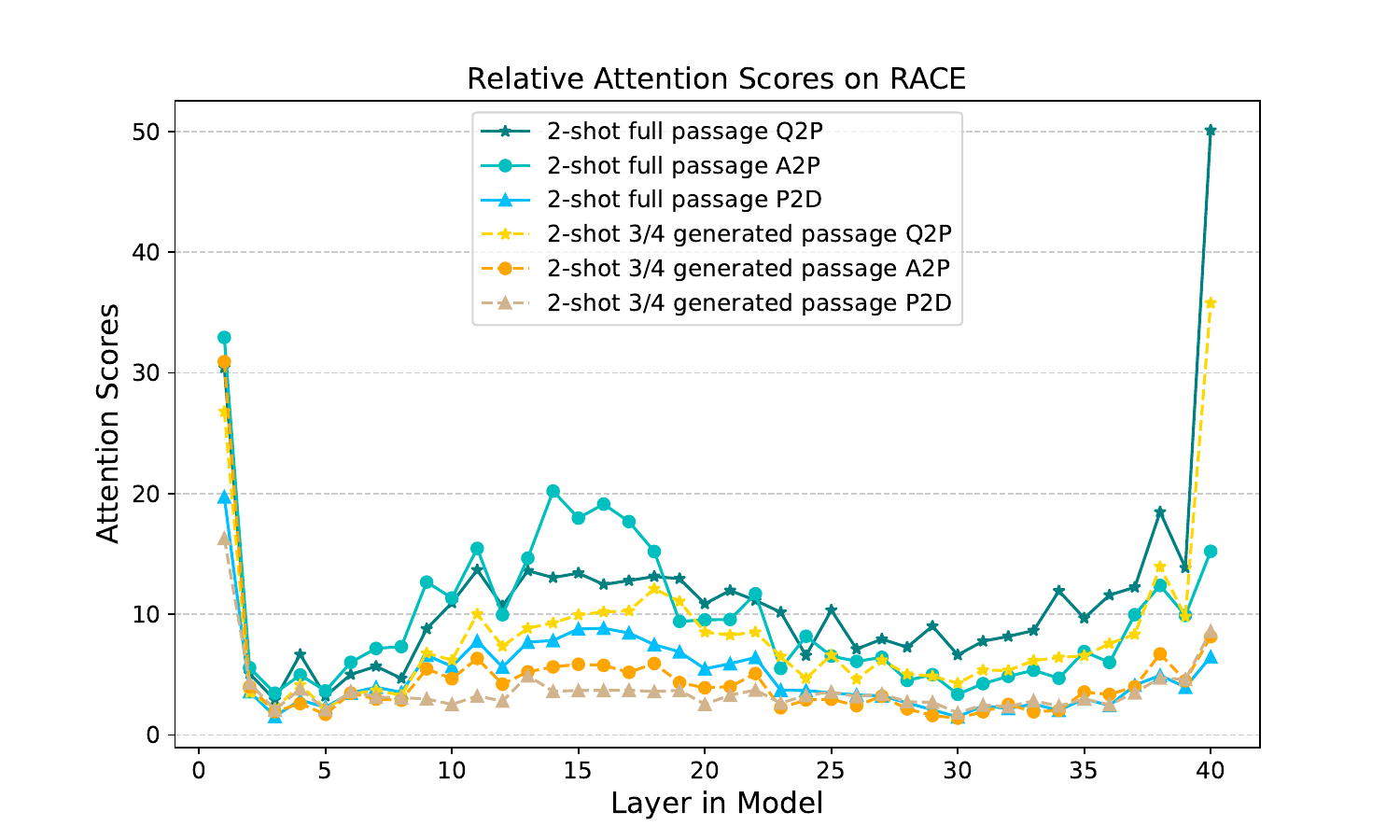} 
\caption{Relative attention scores on RACE with prompts of two settings.
% Layer 1 refers to the first hidden layer in the model.
} % 图片标题
\label{fig/dg_re} % 图片标签，用于引用
\end{figure}

To investigate the underlying reasons for this phenomenon, we visualize the attention scores of the LLM and perform a comparative analysis. The results of RACE are shown in Figure \ref{fig/dg_atten} and Figure \ref{fig/dg_re}. 

Figure \ref{fig/dg_atten} illustrates the impact of two different settings on attention scores: the position of different model layer and different components of prompts. As mentioned in the previous section, the attention score distribution of an input sequence undergoes relatively significant changes as it passes through deeper layers of the model. Initially, the distribution is relatively uniform, but in the middle layers, attention shifts primarily to three parts: the output section within the demonstration, the instruction, and the query. In the attention distribution of the last layer, a trend similar to that of the middle layers can be observed. However, the model shows increased attention to the demonstration compared to the middle layer, probably due to its increased information on overall information in the final layer. Meanwhile, the concentrated attention on the instruction and query sections remains consistent with previous findings. Additionally, the attention distributions in different layers are highly similar between the full Original Passage and the 3/4 Generated Passage.

Figure \ref{fig/dg_re} reveals a similar trend to the previous finding. The experimental setup is similar to that of TriviaQA, However, since the question and the corresponding answer appear before the passage in the demonstration, while the distractors are positioned after the passage. Since the decoder-only architecture only access tokens preceding the current token, the relative attention scores are categorized into three types: Question2Passage, Answer2Passage, and Passage2Distractors. The trend of relative attention scores across layers under both settings is similar to that observed in the QA task. The P2D score is significantly lower than the Q2P and A2P scores, indicating that the connection between the passage and the corresponding target is much weaker than other parts' connection with the passage. When the number of the layers is less than six, the overall attention scores are low, corresponding to a flat attention distribution at the beginning. In deeper layers, the relative attention score and the attention distribution become more directional and focused. Although the trends of the three relative attention scores are generally similar under two settings, the overall relative attention scores for the random generated passage in deeper hidden layers are significantly lower than those for the full passage. This may be because the randomly generated passage has a weaker semantic connection to the corresponding question, answer, and distractors.

\section{Information Flow Results on Distractor Generation}
\label{app:sal}
\begin{figure*}[h] % 开始图片环境
\centering % 使图片居中显示
\includegraphics[width=\textwidth]{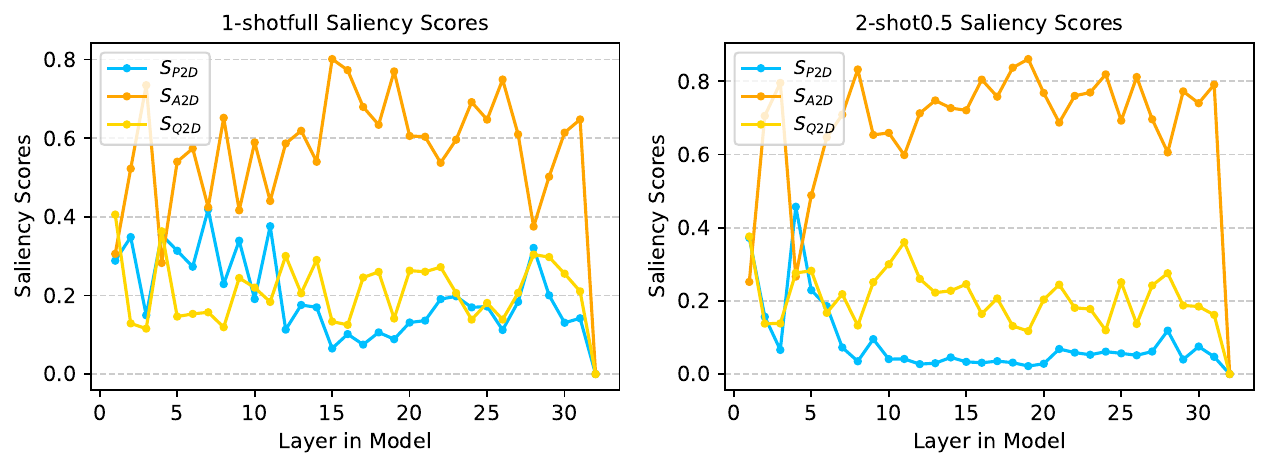} 
\caption{Saliency scores between components of demonstration on RACE.} % 图片标题
\label{fig/saliency_dg} % 图片标签，用于引用
\end{figure*}

In this section, we conduct saliency scoring experiments on RACE, with complete results shown in Figure \ref{fig/saliency_dg}. The $S_{P2D} \text{\&} S_{A2D}$ metrics in Figure \ref{fig/saliency_dg} follow the same definitions as in Section \ref{saliency}, except that the answer is replaced by distractors. As observed in Figure \ref{fig/saliency_dg}, although $S_{P2D}$ reaches relatively higher scores compared to $S_{P2A}$ on TriviaQA, it remains significantly lower than the other metric. This indicates that information primarily flows from answers to distractors, which aligns with our previous findings.
\section{License}
\begin{table}[h]
\renewcommand\arraystretch{1.1}
\centering
\small
\begin{tabular}{ll}
\bottomrule
\textbf{Artifacts} & \textbf{License}\\
\hline
RACE & CMU\\
TriviaQA & Apache-2.0\\
sacreBLEU & Apache-2.0\\
nltk & Apache-2.0\\
Mistral-7B-Instruct-v0.2 & Apache-2.0\\
Llama2-13B-longlora-32k-ft &  Apache-2.0\\
Llama2-13B-Chat & Meta\\
gensim & LGPL-2.1\\
\toprule
\end{tabular}
\caption{Licenses of scientific artifacts we use.}
\label{tab:license}
\end{table}

\end{document}